\documentclass{bmvc2k}

\usepackage{times}
\usepackage{soul}
\usepackage{url}
\usepackage{amssymb}

\usepackage{graphicx}
\usepackage{amsmath}
\usepackage{amsthm}
\usepackage{booktabs}
\usepackage{algorithm}
\usepackage{algorithmic}
\usepackage{subfigure}
\usepackage{multirow}
\usepackage{epsfig}
\usepackage{wrapfig}

\newcommand{\eg}{\emph{e.g.}}

\newcommand{\ie}{\emph{i.e.}}

\makeatletter 
\newcommand\figcaption{\def\@captype{figure}\caption} 
\newcommand\tabcaption{\def\@captype{table}\caption} 
\makeatother

\title{Dynamic Feature Pyramid Networks for Object Detection}

\addauthor{Mingjian Zhu$^{\mbox{\sffamily \scriptsize 1,3,4}}$}{zhumingjian@zju.edu.cn}{1}
\addauthor{Kai Han$^{\mbox{\sffamily \scriptsize 2}}$}{kai.han@huawei.com}{2}
\addauthor{Changbin Yu$^{\mbox{\sffamily \scriptsize 1,3,4}}$}{yu_lab@westlake.edu.cn}{3}
\addauthor{Yunhe Wang$^{\mbox{\sffamily \scriptsize 2}}$}{yunhe.wang@huawei.com}{4}

\addinstitution{$^{\mbox{\sffamily \scriptsize 1}}$Zhejiang University}
\addinstitution{$^{\mbox{\sffamily \scriptsize 2}}$Noah’s Ark Lab\\Huawei Technologies}
\addinstitution{$^{\mbox{\sffamily \scriptsize 3}}$School of Engineering\\ Westlake University}
\addinstitution{$^{\mbox{\sffamily \scriptsize 4}}$Institute of Advanced Technology\\ Westlake Institute for Advanced Study}

\runninghead{ZHU ET AL.}{Dynamic Feature Pyramid Networks for Object Detection}
\def\eg{\emph{e.g}\bmvaOneDot}

\begin{document}
	
	\maketitle
	
	\begin{abstract}
		Feature pyramid network (FPN) is a critical component in modern object detection frameworks. The performance gain in most of the existing FPN variants is mainly attributed to the increase of computational burden. An attempt to enhance the FPN is enriching the spatial information by expanding the receptive fields, which is promising to largely improve the detection accuracy. In this paper, we first investigate how expanding the receptive fields affect the accuracy and computational costs of FPN. We explore a baseline model called inception FPN in which each lateral connection contains convolution filters with different kernel sizes. Moreover, we point out that not all objects need such a complicated calculation and propose a new dynamic FPN (DyFPN). The output features of DyFPN will be calculated by using the adaptively selected branch according to a dynamic gating operation. Therefore, the proposed method can provide a more efficient dynamic inference for achieving a better trade-off between accuracy and computational cost. Extensive experiments conducted on MS-COCO benchmark demonstrate that the proposed DyFPN significantly improves performance with the optimal allocation of computation resources. For instance, replacing inception FPN with DyFPN reduces about 40\% of its FLOPs while maintaining similar high performance.
	\end{abstract}
	
	\section{Introduction}
	\label{sec:intro}
	Object detection~\cite{lin2017feature,song2020fine,guo2020augfpn} is a fundamental task in the computer vision field, which attracts growing attention in recent years. A practical method to detect objects precisely can be useful in modern applications, such as surveillance video and robot navigation. Recent progress in object detection largely stems from the exploitation of deep convolutional neural network (CNN). Devising an effective CNN-based architecture is the mainstream approach for detecting objects across a wide range of scales. The modern detection framework can be categorized as one-stage approach and two-stage approach. The one-stage approaches such as YOLO~\cite{redmon2016you}, SSD~\cite{liu2016ssd}, FCOS~\cite{tian2019fcos} and CenterNet~\cite{duan2019centernet}, directly extract features to predict object classes and locations. In contrast, the two-stage approaches, \eg Faster R-CNN~\cite{ren2015faster}, and Cascade R-CNN~\cite{cai2018cascade}, firstly obtain the region of interests (ROI) by region proposal network and further generate refined bounding boxes and classes based on ROI. Both approaches make great progress in recent years.
	
	Currently, many methods~\cite{li2019scale,dai2017deformable,peng2019pod} expands the receptive fields for multi-scale feature learning and achieve encouraging results. For example, RFB~\cite{liu2018receptive} replaces the later convolution layers of SSD~\cite{liu2016ssd} with a multibranch convolutional block to enhance the features in multiple scales. 
	Although these methods have made tremendous efforts for expanding the receptive fields of their detectors, how different kinds of convolutional kernels could affect FPN has not been fully investigated. To this end, we conduct extensive experiments to discuss the benefits and deficiencies brought by different combinations of convolutions. 
	
\begin{figure*}[tb]
		\centering
		\subfigure[inception FPN]{\label{Inception FPN}
			\includegraphics[height=3.2cm]{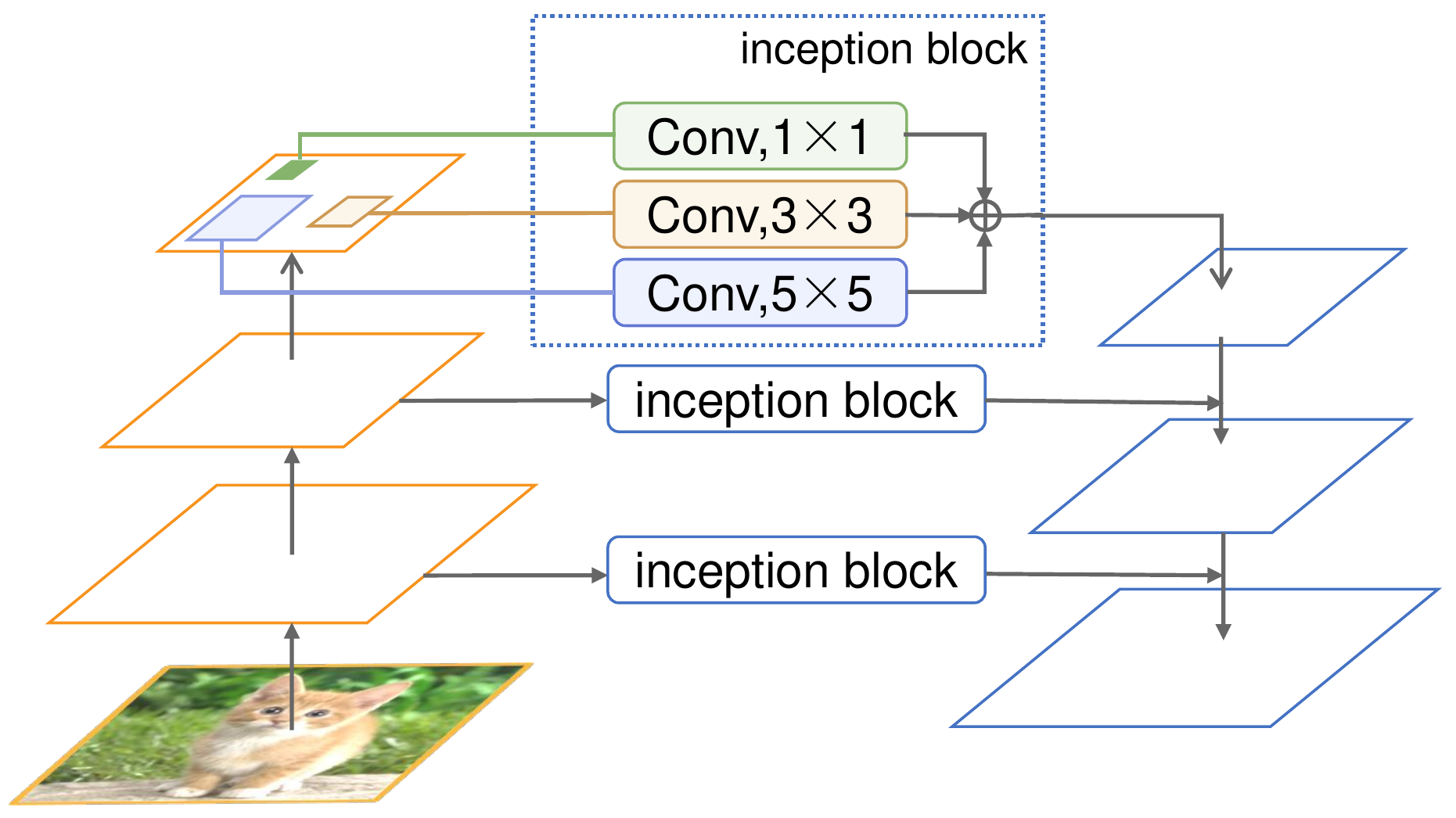}}
		\quad
		\subfigure[DyFPN]{\label{Dynamic FPN structure b}
			\includegraphics[height=3.2cm]{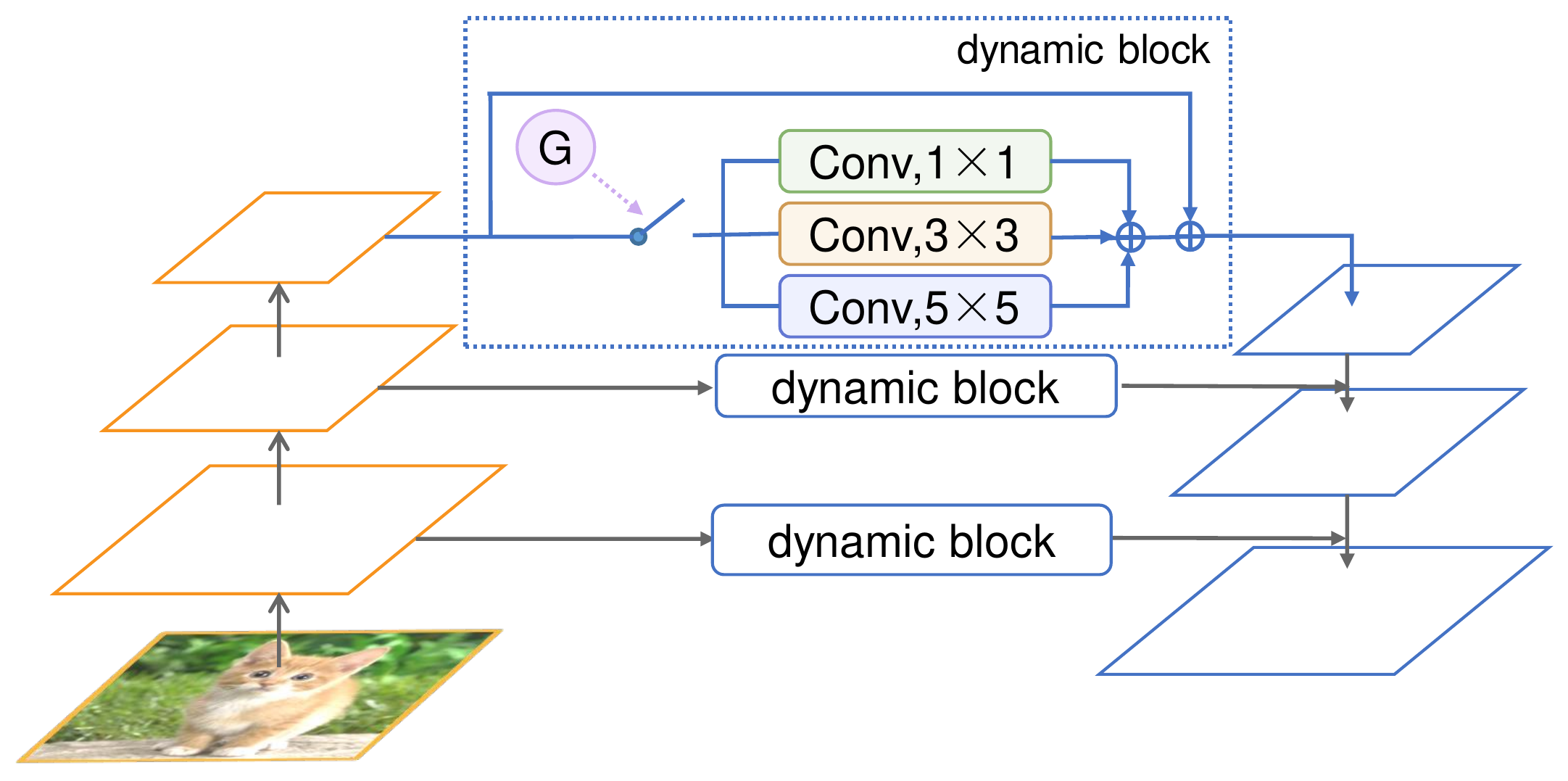}}
		\quad
		\caption{The comparison among the inception FPN and the DyFPN. The content in the dashed box represents the details of the inception block and dynamic block. The "G" denotes our proposed dynamic gate.}
		\label{fig:Inception and Dynamic network}
	\end{figure*}

	
	To explore the representation ability of features generated by different convolutions, we first embed the inception blocks into the conventional FPN. As shown in Figure~\ref{Inception FPN}, each inception block in the investigated inception FPN contains convolution filters with different kernel sizes (\eg, $1\times 1$, $3\times 3$, $ 5\times 5$). Better features can be effectively generated by a combination of different convolutions. Compared with the conventional FPN, the inception FPN enriches the spatial information by fully expanding the receptive fields, which significantly improves the detection accuracy.

	
	
	%
	
	However, the combination of filters with different kernel sizes obviously increases the overall computational costs. In addition, the difficulties for detecting objects in different natural images are exactly variant, which implies that not all objects need such a complicated computation for correct prediction. These observations further motivate us to introduce the dynamic mechanism~\cite{li2020learning,wang2018skipnet} into the inception FPN to balance the performance and the overall computation burden. As shown in Figure~\ref{Dynamic FPN structure b}, a learnable dynamic gate with negligible computational costs is inserted before an inception block in each lateral connection. The dynamic gate adaptively determines whether to execute the whole inception block based on the input. We perform experiments to show that the introduced dynamic gate can largely reduce the computational cost of the baseline model (\ie, the inception FPN) while maintaining similar high detection accuracy. 
	
	In this paper, we first perform extensive experiments to investigate the impact of applying combinations of different convolutions on FPN. Although utilizing these convolutions improves the detection accuracy by enriching the spatial information, it also leads to a significant computational burden. Thus, we further propose DyFPN to overcome this problem. DyFPN adaptively determines whether to conduct the multiple convolutions based on the input images. The effectiveness of DyFPN is well evaluated on various backbone architectures. Experimental results on the MS-COCO benchmark show that replacing the baseline inception FPN with DyFPN consistently saves significant computational costs while preserving similar high accuracy.
	
	We organize the rest of the paper as follows. In Section~\ref{Related Work}, we investigate the related works on object detection and dynamic neural networks. Section~\ref{Methods} demonstrates the details of the investigated baseline inception FPN and our proposed DyFPN. Section~\ref{experiments} evaluates the inception FPN and DyFPN on benchmark datasets and various backbone architectures and Section~\ref{Conclusion} concludes this paper.

	\section{Related Work}
	\label{Related Work}
	\paragraph{Object Detection.} The objection Detection task aims at recognizing what the object is and where the object locates in an image. Benefit from the deep neural network, many methods~\cite{girshick2014rich,redmon2016you,lin2017focal} on object detection tasks have achieved impressive improvements in recent years. Faster R-CNN~\cite{ren2015faster} proposes an end-to-end detection approach by replacing Selective Search with a novel region proposal network. SSD~\cite{liu2016ssd} predicts a series of bounding boxes with different scales and aspect ratios from several feature layers. Cascade R-CNN~\cite{cai2018cascade} trains the model in a cascade manner with gradually increased IoU threshold. Exploiting features from different scales have been verified effective in detection task~\cite{liu2016ssd,pang2019libra,wang2019carafe,guo2020augfpn}. FPN~\cite{lin2017feature} constructs a bottom-up pathway, a top-down pathway, and lateral connections to fuse the features with different resolutions and scales in an efficient way. CARAFE~\cite{wang2019carafe} proposes an effective feature upsampling operator and integrates it into FPN to boost the performance. AugFPN~\cite{guo2020augfpn} refines the FPN using Consistent Supervision, Residual Feature Augmentation, and Soft RoI Selection, simultaneously. Libra R-CNN~\cite{pang2019libra} strengthens the multi-level features in the feature pyramid by using the integrated balanced semantic features. 
	\paragraph{Dynamic Neural Networks.} 
	Previous works on dynamic neural networks mainly focus on adjusting the architecture of models according to the input images~\cite{li2020learning,wang2018skipnet,huang2017multi,yuan2019s2dnas,wu2018blockdrop}. MSDNet~\cite{huang2017multi} adopts dynamic evaluation, which tackles easy examples at early classifiers while handles hard examples with the whole network. S2DNAS~\cite{yuan2019s2dnas} proposes a method to transform various CNN models into dynamic models without manually re-designing. For image classification, ConvNet-AIG~\cite{Veit2018} designs gates to determine whether to execute or skip the specific layers, which enables the dynamic adjustment of inference graphs conditioned on the input features. Dynamic routing~\cite{li2020learning} searches scale transform paths on the fly for semantic segmentation. Different from the previous dynamic networks, our DyFPN aims at alleviating the computational costs caused by the combinations of different convolutions while still benefiting from enriched spatial information.

	\vspace{1cm}
	
	\section{Methods}
	\label{Methods}
	
	\subsection{Inception FPN}
	\label{The Basic Building Block}
	By effectively leveraging features from different layers in the backbone model, the feature fusion method is widely used to improve the network performance~\cite{wang2019learning,kim2018parallel,lin2017feature}. An efficient method to fuse features is building a feature pyramid. Typically, for a list of input features with different scales, the FPN takes a series of features $\{F_2, F_3, F_4, F_5\}$ as input and outputs the aggregated features $\{P_2, P_3, P_4, P_5\}$ as follows:
	\begin{align}
		P_{5} = & f_5(F_5),\\
		P_l = & f_l(F_l) + R(P_{l+1}), \ l=2,3,4,
	\end{align}
	where $l$ denotes the level of the pyramid. $R$ denotes the resizing operation to generate the features that are respectively of the same spatial sizes. The lateral connection $f_l(\cdot)$ is typically a 1$\times$1 convolutional layer which lacks sufficient spatial information for recognizing objects. To this end, we investigate an intuitive model called inception FPN, which enriches the spatial information of the feature pyramid by expanding the receptive fields. The inception FPN utilizes the inception block in the lateral connection and achieves significant advances in detection accuracy. As shown in Figure~\ref{Inception FPN}, the inception block consists of a set of convolutions with kernel sizes of 1$\times$1, 3$\times$3, and 5$\times$5 and sums the features from different convolutions as follows:
	\begin{equation}
		f_l(F_l) = \textit{Conv}_{1\times1}(F_l) + \textit{Conv}_{3\times3}(F_l) + \textit{Conv}_{5\times5}(F_l),
	\end{equation}
	where $l=2,3,4,5$. With convolutions of different kernel sizes in the lateral connection, the extracted features at each level benefit from different receptive fields. In the section~\ref{experiments}, we conduct extensive experiments to demonstrate that gradually replacing the 1$\times$1 convolutions with the more convolutions (\ie, 1$\times$1, 3$\times$3, 5$\times$5, and their dilated variants) can obtain richer spatial information and achieve better performance.

	\subsection{Dynamic Feature Pyramid Network}
	\label{Dynamic FPN}
	The inception FPN can largely improve the detection accuracy but brings heavy computational burdens. To this end, we propose DyFPN, which aims at tackling the problems of inception FPN by introducing a novel dynamic block. Basically, the dynamic block consists of three components: dynamic gate, inception block, and skip-connection. The experiments in section~\ref{experiments} demonstrate that the combination of these components achieves a better trade-off between accuracy and computational cost when compared to the inception FPN.

	
	\paragraph{Dynamic Block}
	In inception FPN, the lateral connections are static, which means it executes the same convolutions in the inference stage. However, it is intuitive that the detection difficulties of various input images are different, which implies that some ''easy'' images can be detected correctly without the enriched spatial information. Thus, the inception FPN contains computation redundancy. In contrast, our proposed dynamic block in DyFPN adaptively determines the operations in lateral connections based on the input image. The dynamic architecture for a specific image can largely reduce the computational cost and retain high detection accuracy. Figure~\ref{Dynamic FPN structure b} shows the details of the dynamic block. The motivation of the dynamic gate aims at predicting a one-hot vector, which denotes whether to execute or skip the branch of the inception block. We sum the features from a set of convolutional layers in the inception block. We always conduct a $1 \times 1$ convolutional layer at each level and consider it as skip-connection. The features from the branch of the inception block and the skip-connection are fused by summation. We insert the dynamic blocks to all the lateral connections of the feature pyramid. In training, the prediction of the gate is multiplied by the aggregated features from the dynamic block. In testing, the dynamic block does not need to be executed if the gate predicts 0 (\ie, the decision of skipping the inception block).
	
	In the dynamic gate, we first apply a non-linear function $g_l(\cdot)$ on $F_l$ and produces the logits of the gate signals:
	\begin{equation}
		\alpha_l = g_l(F_l),
	\end{equation}
	where $a_l\in\mathbb{R}$ which determines the sampling probability of inception block at level-$l$. Then the one-hot gate vector $\beta_l$ is obtained by Gumbel Softmax function as follows:
	\begin{align}
		\begin{split}
			\beta_{l}^i &= \textit{GumbelSoftmax}(\alpha_l^i|\alpha_{l})\\
			&= \frac{exp[\alpha_{l}^{i}+n_{l}^{i}/\tau]}{\sum_{i}exp[(\alpha_{l}^{i}+n_{l}^{i}/\tau)]}
		\end{split}
		\label{eq:gumbelsoftmax}
	\end{align}
	where $\beta_l^i\in\{0,1\}$, and $n_{l}^{i} \sim \textit{Gumbel}(0,1)$ is a random noise sampled from the Gumbel distribution. $\tau$ is a temperature parameter that influences the Gumbel Softmax function. Specifically, inspired by squeeze-and-excitation (SE) module~\cite{hu2018senet}, our dynamic gate is composed of a global average pooling layer, two fully-connected layers, and a ReLU layer:
	
	\begin{equation}
		\label{FCGate}
		\alpha_l = W_{2}(\delta(W_{1}P(F_l)+b_1))+b_2,
	\end{equation}
	where $P$ denotes the global average pooling layer. $\delta$ means the ReLU activation function. The input feature with a shape of $(C, H, W)$ from the selected layer in the backbone model is firstly squeezed by an average pooling operation in each channel to produce the feature of size $C$. Then we reduce the feature dimension by 4 in the first fully-connected layer. The second fully-connected layer, a non-linear activation function, and a Gumbel Softmax function~\cite{Wu2019FBNetHE} are further leveraged to generate the one-hot vector for the dynamic block. The introduced pooling operation largely reduces the computational costs of fully-connected layer applied on the input features, which makes the computational burden of the dynamic gate can be negligible.


	\paragraph{Resource Constraint.}In real-world scenarios, physical devices often impose different computing resource constraints on the model. Thus, DyFPN should be developed with the consideration of different computational expenses. However, if we use the detection loss only, the dynamic gate will tend to provide a sub-optimal solution, which conducts as many inception blocks as possible because features with the most enriched spatial information correspond to relatively lower detection loss generally. In order to achieve a satisfactory efficiency-accuracy trade-off, we propose a new loss $\mathcal{L}_{C}$ to guide the training. Here, we denote $C$ as the computational cost of the model, \ie, FLOPs. The maximum and minimum costs of the dynamic block can be calculated before training the model and we denote them as $C_{max}$ and $C_{min}$. For DyFPN, the decision of executing or skipping the inception block in all lateral connections of the feature pyramid leads to maximum or minimum computational cost, respectively. Here we introduce the losses for an end-to-end optimization:
	\begin{align}
		\begin{split}
			\mathcal{L}_{C} &= ((C_R - C_{target})/(C_{max} - C_{min}))^2\\
		\end{split}
	\end{align}
	where $C_R$ denotes the real computational cost of the dynamic block. $C_{target}$ represents the target resource cost. We can control the target cost by setting the hyper-parameter $\alpha\in(0, 1)$. The target cost can be formulated as follows:
	\begin{align}
		\begin{split}
			C_{target} &= C_{min} + \alpha * (C_{max} - C_{min})\\
		\end{split}
	\end{align}
	The total loss function can be optimized as follows:
	\begin{align}
		\begin{split}
			\mathcal{L} &= \mathcal{L}_{Det} + \lambda\mathcal{L}_{C}\\
		\end{split}
	\end{align}
	where $\mathcal{L}_{Det}$ and $\mathcal{L}_{C}$ represent the loss of detection and computational cost, respectively. We leverage $\lambda$ to balance the detection accuracy expectation and computational cost constraints, respectively.

	\section{Experiments}
	\label{experiments}
	\subsection{Dataset and Evaluation Metrics}
	All experiments are conducted on the MS-COCO 2017 detection dataset~\cite{Lin2014MicrosoftCC}, which contains 80 object categories. Following the protocol in MS-COCO, 118k images are used for training. We report the results of ablation studies for \textit{minival} with 5k images. Following common practice, MS-COCO Average Precision(AP) with different IoU thresholds is leveraged as the evaluation metric~\cite{guo2020hit,zhao2019m2det}. 
	
	\subsection{Implementation Details}
	For all experiments, our detectors are trained end-to-end on a machine with 4 NVIDIA RTX 2080Ti GPUs. We utilize the SGD optimizer to train the model for 12 epochs, known as $1 \times$ schedule. Linear warm-up strategy for 500 iterations is leveraged at the beginning of training. We initialize the learning rate as 0.01 and decrease it to 0.001 and 0.0001 at 8th-epoch and 11th-epoch. The momentum is set as 0.9 and the weight decay is 0.0001. The batch size is set to 2 per GPU. The input size is $1333 \times 800$. The factor $\tau$ in Eq.~\ref{eq:gumbelsoftmax} is set to 1.0. Our implementation is developed by PyTorch framework~\cite{paszke2019pytorch} and mmdetection toolbox~\cite{mmdetection}.

	\subsection{Main Results}
	
	\paragraph{Effectiveness and Efficiency of DyFPN.}
	We compare DyFPN with model variants of FPN and inception FPN in Figure~\ref{fig:FPN_Inception_DYFPN_FLOPS_mAP}. We can see that the DyFPN achieves the best efficiency-accuracy trade-off among the model variants. In each lateral connection of the FPN, we replace $1 \times 1$ convolutional layer with $3 \times 3$, $5 \times 5$ and $7 \times 7$ convolutional layer, respectively. The FPN with a $7 \times 7$ convolution filter achieves the best performance while consumes the most computational costs. The model with the lowest accuracy follows the architecture of the conventional FPN. It can be observed that enlarging the size of the convolution filter in the FPN improves the accuracy while consumes more computational costs, which is not efficient. We combine different kinds of convolution filters and their dilated variants in the inception FPN, which accounts for different computational costs and accuracies. Although simply adding more convolution filters benefits the accuracy, the consequent computational costs drastically increase. To this end, the inception FPN is inefficient. Our DyFPN always outperforms the FPN and the inception FPN with similar computational costs. From the comparisons, we can conclude that our proposed method can largely improve the performance of the FPN in an efficient manner.

	\paragraph{Extension to Different Backbones.}
	We compare the DyFPN with the inception FPN on different backbones to demonstrate the effectiveness of DyFPN. In Table~\ref{tab:comparison_inceptionFPN_DyFPN}, we can see that the DyFPN largely saves the computational costs of the baseline with negligible performance variation on all backbones. Based on Faster R-CNN, the DyFPN (537.5G) reduces up to 40\% computational costs of the inception FPN (896.2G) with Resnet-50 as the CNN backbone, and the accuracy only drops 0.2 AP. The accuracy of DyFPN on Cascade R-CNN with ResNet-101 is even better than that of the inception FPN and the FLOPs also decreases 37.1\%. Basically, the methods based on Cascade R-CNN~\cite{cai2018cascade} detects more accurately than that based on Faster R-CNN, but Faster R-CNN saves more computational costs. The DyFPN achieves the best results with Cascade R-CNN and Resnet-101 as the backbone. Compared with the inception FPN, the computational costs of DyFPN are basically reduced by 34\%-40\% while the accuracy only drops 0.2-0.3 AP on most backbones.

	\begin{table}[tp]	
		\begin{minipage}[b]{0.45\linewidth}
			
			\includegraphics[width=1.0\linewidth]{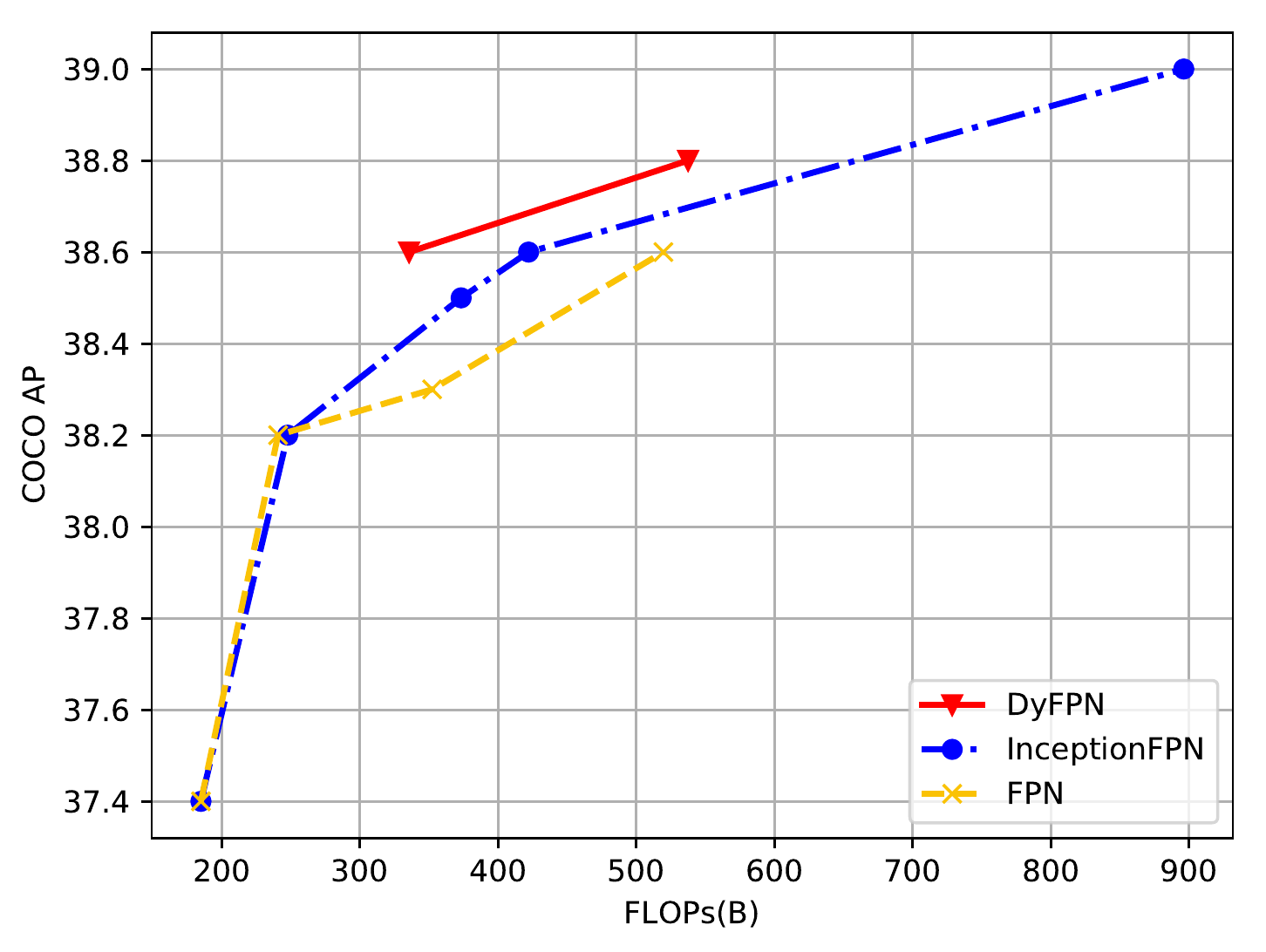}
			\hspace{-0.9cm}		
			\figcaption{Accuracy \textit{v.s.} FLOPs. Faster R-CNN with ResNet-50 is utilized as the backbone.}
			\label{fig:FPN_Inception_DYFPN_FLOPS_mAP}
		\end{minipage}
		\hspace{1mm}
		\begin{minipage}[b]{0.50\linewidth}
			\footnotesize
			\begin{tabular}{c|c|c|c}
				\hline
				Backbone                                                                            & Model         & AP   & \#FLOPs(G) \\ \hline
				\multirow{2}{*}{\begin{tabular}[c]{@{}c@{}}Faster R-CNN\\ ResNet-50\end{tabular}}   & Baseline & 39.0 & 896.2      \\
				& DyFPN         & 38.8 & 537.5 ($\downarrow40.0\%$)     \\ \hline
				\multirow{2}{*}{\begin{tabular}[c]{@{}c@{}}Faster R-CNN\\ ResNet-101\end{tabular}}  & Baseline & 40.5 & 963.6      \\
				& DyFPN         & 40.4 & 592.4 ($\downarrow38.5\%$)     \\ \hline
				\multirow{2}{*}{\begin{tabular}[c]{@{}c@{}}Cascade R-CNN\\ ResNet-50\end{tabular}}  & Baseline & 41.3 & 923.9      \\
				& DyFPN         & 40.9 & 561.0 ($\downarrow39.3\%$)     \\ \hline
				\multirow{2}{*}{\begin{tabular}[c]{@{}c@{}}Cascade R-CNN\\ ResNet-101\end{tabular}} & Baseline & 42.6 & 991.2      \\
				& DyFPN         & 42.7 & 623.1 ($\downarrow37.1\%$)     \\ \hline
				\multirow{2}{*}{\begin{tabular}[c]{@{}c@{}}Mask R-CNN\\ ResNet-50\end{tabular}}     & Baseline & 39.5 & 915.8      \\
				& DyFPN         & 39.2 & 587.0 ($\downarrow36.9\%$)     \\ \hline
				\multirow{2}{*}{\begin{tabular}[c]{@{}c@{}}Mask R-CNN\\ ResNet-101\end{tabular}}    & Baseline & 41.0 & 981.9      \\
				& DyFPN         & 40.9 & 644.5 ($\downarrow34.4\%$)     \\ \hline
			\end{tabular}
			\tabcaption{The comparison of the baseline and DyFPN on different backbones. The baseline denotes the inception FPN.}
			\label{tab:comparison_inceptionFPN_DyFPN}
		\end{minipage}
	\end{table}

	\subsection{Ablation Study}

	\paragraph{Inception FPN.}
	In Table~\ref{tab:different convolution layers}, We conduct experiments to demonstrate that the enriched spatial information and expanded receptive fields significantly improve detection performance. We can see that gradually aggregating more features from different convolutions leads to accuracy improvement. We first follow the architecture of FPN and demonstrate its performance in the setting of \textit{k}=1, \textit{d}=1, where \textit{k} denotes the kernel size and \textit{d} denotes the dilation rate. Based on FPN, we add ${3 \times 3}$ convolution in each lateral connection, which leads to 0.8 absolute gain in AP. We can see that the addition of ${3 \times 3}$ convolution brings additional improvements for all sizes of objects. We further add ${5 \times 5}$ convolution and find a 0.4 increase in AP. However, the addition of ${5 \times 5}$ convolution achieves improvement for middle and large objects but drops in performance for small objects. We conjecture that the receptive field of ${5 \times 5}$ convolution overemphasizes large objects, which drowns the information of the small objects in the aggregated features. We add some convolutions with different dilation rates to further increase the receptive field. Note that although the ${3 \times 3}$ convolution with dilation rate 2 has the same receptive field as the ${5 \times 5}$ convolution, their parameters and computations are not the same. Thus, employing \textit{k}=3, \textit{d}=2 or \textit{k}=5, \textit{d}=1 leads to different results in the inception block. It can be observed that the addition of convolutions with a dilation rate of 2 and 3 improves the detection accuracy in almost all scales. Aggregating features from convolutions of different kernel sizes and dilation rates enables the receptive fields of the detection model to be expanded and the spatial information can be enriched, which results in an accuracy improvement. The method in the last row benefits from all kinds of convolutions and achieves the best results. To this end, the convolutional layers in our inception block finally adopt the following configurations: kernel size = [1, 3, 3, 3, 5, 5, 5], dilation rate = [1, 1, 2, 3, 1, 2, 3], padding = [0, 1, 2, 3, 2, 4, 6]. 
	
	In Figure~\ref{fig:feature_visulization}, we visualize the features generated by convolutions in inception FPN. We can see that simply exploring the spatial information in the feature pyramid cannot obtain satisfactory results while utilizing the convolutions with different kernel sizes can generate enriched spatial features for detection.
	
	\begin{table*}[tb]
		\centering
		\small
		\begin{tabular}{cccccccccccccc}
			\hline
			\begin{tabular}[c]{@{}l@{}}k=1 \\ d=1\end{tabular} & \begin{tabular}[c]{@{}l@{}}k=3 \\ d=1\end{tabular} & \begin{tabular}[c]{@{}l@{}}k=5 \\ d=1\end{tabular} & \begin{tabular}[c]{@{}l@{}}k=3 \\ d=2\end{tabular} & \begin{tabular}[c]{@{}c@{}}k=3\\  d=3\end{tabular} & \begin{tabular}[c]{@{}c@{}}k=5\\  d=2\end{tabular} & \multicolumn{1}{c|}{\begin{tabular}[c]{@{}c@{}}k=5\\ d=3\end{tabular}} & AP & AP$_{50}$ & AP$_{75}$ & AP$_{S}$ & AP$_{M}$ & AP$_{L}$ \\ \hline
			\checkmark      &       &       &         &        &      & \multicolumn{1}{l|}{}   &{37.4} &{58.3} &{40.4} &{21.2} &{41.1} &{48.6}        \\
			
			\checkmark  &\checkmark       &       &       &      &          & \multicolumn{1}{l|}{}   &{38.2} &{59.2} &{41.1} &{22.1} &{41.8} &{49.4}   \\
			\checkmark      &\checkmark        &\checkmark        &       &   & & \multicolumn{1}{l|}{}    &{38.6} &{59.9} &{41.6} &{21.9} &{42.1} &{50.4}        \\
			\checkmark  &\checkmark &     &\checkmark     &\checkmark       &      & \multicolumn{1}{l|}{}     &{38.5} &{59.9} &{41.8} &{22.0} &{42.2} &{50.1}  \\
			\checkmark  &\checkmark  &\checkmark  &\checkmark  &\checkmark &\checkmark   &\multicolumn{1}{c|}{\checkmark}  &\textbf{39.0} &\textbf{60.6} &\textbf{42.4} &\textbf{23.3} &\textbf{42.7} &\textbf{50.5} \\	
			\hline 
		\end{tabular}
		\caption{Ablation study of inception FPN. The Faster R-CNN framework with ResNet-50 model is utilized as the backbone. The best scores are \textbf{bold}.}
		\label{tab:different convolution layers}
	\end{table*}

	\begin{table}[tp]	
		\begin{minipage}[b]{0.51\linewidth}
			\hspace{-2.5mm}
			\includegraphics[width=1.0\linewidth]{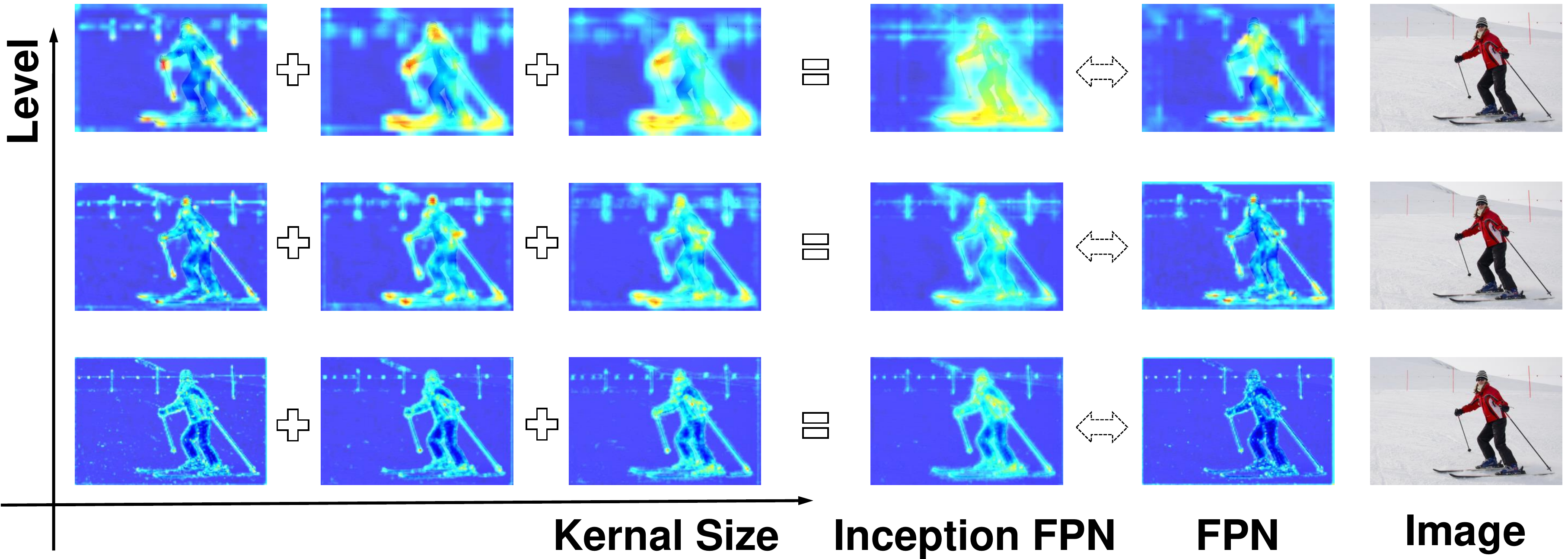}
			\figcaption{The visulization of the features.}	
			\centering	
			\label{fig:feature_visulization}
		\end{minipage}
		\begin{minipage}[b]{0.47\linewidth}
			\label{tab:Comparison between Dynamic FPN and FPN with Random Gate}	
			\footnotesize
			\begin{tabular}{l|c|c|c|c}
				\hline
				{Model}  &$\alpha$ &$\lambda$ &AP &\#FLOPs(G)  \\ 
				\hline
				{Baseline}  &{-} &{-} &{39.0} &{896.2}\\		
				{DyFPN-Raw}  &{0.0} &{0.0} &{38.8} &537.5\\
				{DyFPN-I}  &{0.5} &{0.1} &{38.7} &523.6\\
				
				{DyFPN-II}  &{0.3} &{0.1} &{38.6} &399.3\\
				
				{DyFPN-III}  &{0.2} &{0.1} &{38.4} &329.0\\
				
				{DyFPN-IV}  &{0.2} &{0.5} &{38.3} &328.6\\
				\hline
			\end{tabular}%
			
			\tabcaption{Ablation studies of different resource budgets.}
			\label{tab:resource budgets}
		\end{minipage}
	\end{table}

	%

	\paragraph{Computational Resource Budgets.} To obtain a satisfactory efficiency-accuracy trade-off, we propose a computational cost loss to restrict the computational cost of the model. By tuning the hyper-parameter $\alpha$ and $\lambda$, we train DyFPN variants with different resource constraints, as shown in Table~\ref{tab:resource budgets}. We compare the resource constraint models with DyFPN-Raw, which is trained without the constraint. By tuning $\alpha$, we can set different target computational costs for DyFPN. It can be seen that the computational costs of DyFPN descend to different levels with different targets. With the strongest constraint, the FLOPs of DyFPN-IV is decreased to 61.1\% of DyFPN-Raw. Besides, the computational loss enables DyFPN-I to consume fewer computational costs while preserving similar accuracy as DyFPN-Raw.

	\paragraph{Realistic Acceleration.}
	Based on the Faster R-CNN framework and ResNet-50 model, we compute the latency, FLOPs, and AP for inception FPN and DyFPN in Table~\ref{tab:latency}. Considering the data-dependent property of the DyFPN, we report the average latency and FLOPs in the MS-COCO benchmark. We can see that the DyFPN still largely reduces the forward time in the realistic calculation without any special software or hardware designs. 
	
	\paragraph{Effectiveness of the Dynamic Gate.}
	
	%

	To demonstrate the effectiveness of our dynamic gate, we compare its performance with a random number generator. In Table~\ref{tab:Comparison between Dynamic FPN and FPN with Random Gate}, we apply different decisions to DyFPN in the training and testing stage. Here, \textit{R} denotes the random decisions and \textit{G} denotes the decisions from our dynamic gate. The backbone is Faster R-CNN with ResNet-50. The random decisions are randomly generated one-hot vectors. we can see that replacing the gate decisions with the randomly generated decisions in the testing period always leads to a dramatic performance drop, which indicates that the gate decisions are significant. For a fair comparison, we test the model 10 times with random decisions and average their results. Considering that the inconsistency between train/test-time inference leads to performance degradation to the model, we also train the models with random decisions. The results show that training the DyFPN with random decisions makes the model robust to the disturbance of the stochastic factors in testing. However, our gate decisions still outperform the random decisions with a large margin.

	\begin{table}[tp]	
		\begin{minipage}[h]{0.58\linewidth}
			\small
			\begin{tabular}{l|c|c|c}
				\hline
				{Model}  &Latency(ms) &\#FLOPs(G) &AP \\ 
				\hline
				{Baseline}  &{287.6} &{896.2} &{39.0}\\
				{DyFPN}  &{188.6} ($\downarrow34.4\%$) &{537.5} ($\downarrow40.0\%$) &{38.8} \\
				\hline
			\end{tabular}%
			\tabcaption{Comparison of the latency, the computational costs, and the accuracy of different models on a RTX 2080Ti GPU. }
			\label{tab:latency}
		\end{minipage}
		\hspace{0.5mm}
		\begin{minipage}[h]{0.40\linewidth}
			\small
			\begin{tabular}{c|c|c|c|c}
				\hline
				\centering	
				{Train}  &Test &AP &AP$_{50}$  &AP$_{75}$ \\ 
				\hline
				{R}  &{R} &36.9   &57.7 &39.9 \\
				{G}  &{R} &18.8   &30.0 &19.8 \\
				{G}  &{G} &38.8   &60.5 &42.0 \\
				\hline
			\end{tabular}%
			\tabcaption{Ablation studies of different decision sources.}
			\label{tab:Comparison between Dynamic FPN and FPN with Random Gate}
		\end{minipage}
	\end{table}	
	
	\begin{figure}[t]
		\centering
		\includegraphics[height=0.46\linewidth]{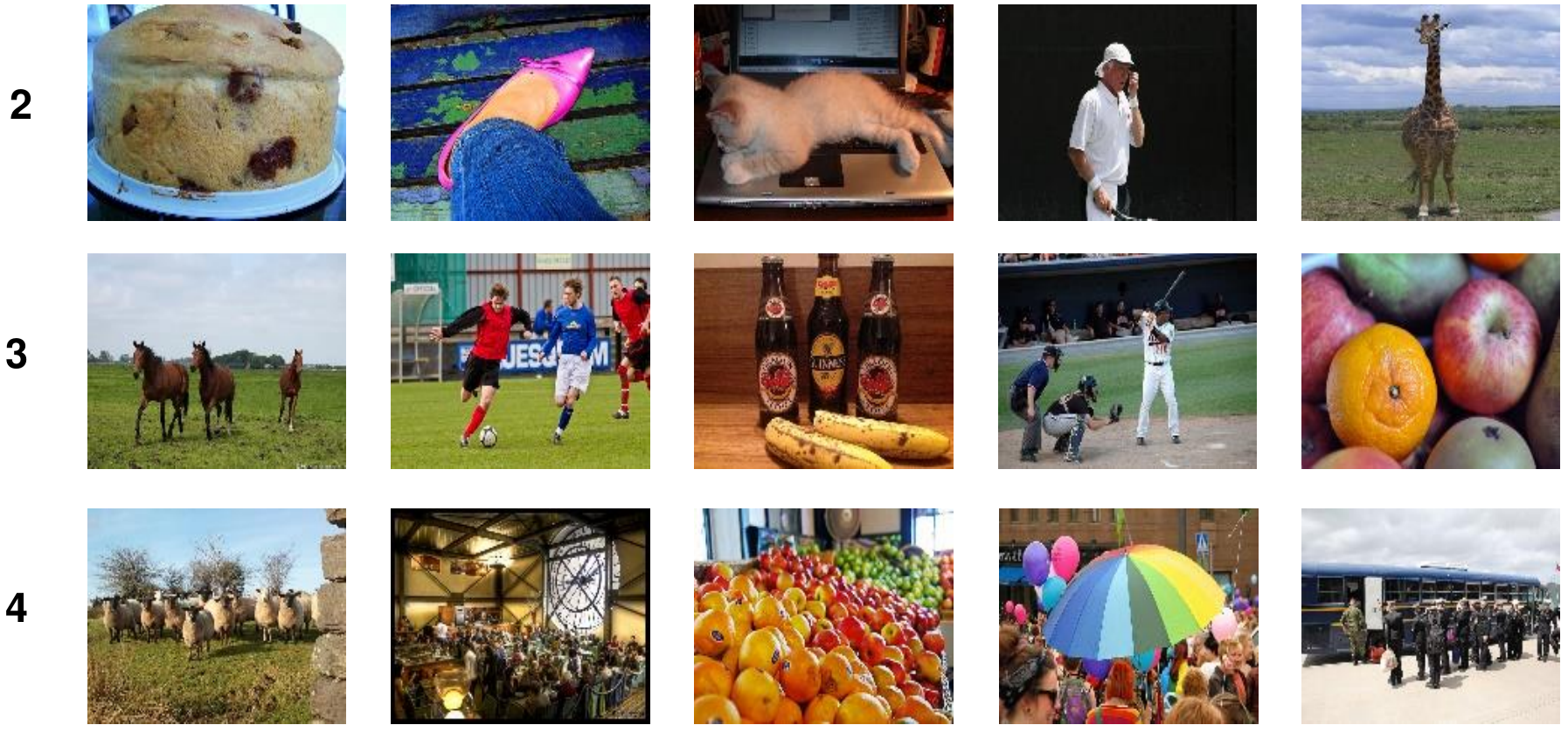}
		\caption{Qualitative results of the DyFPN. Each value denotes the number of the executed inception blocks for these images.}
		\label{fig:different number of executed dynamic blocks}
	\end{figure}

	\paragraph{Number of the Executed Inception Blocks in DyFPN.}
	We demonstrate the decisions of the dynamic gates and their corresponding images in DyFPN as shown in Figure~\ref{fig:different number of executed dynamic blocks}. There are four lateral connections in DyFPN. At each connection, the gate determines whether to execute the inception block. For each input, we count the number of executed blocks in all connections of DyFPN. From the visualization results, the number of executed blocks generally increases with the complexity of the objects, which is compatible with the human perception system~\cite{walther2011simple}. For example, in the first row, the number of objects is small and the target occupies most of the image. Thus, two inception blocks in DyFPN are sufficient to give correct predictions. The middle five images which contain more objects are decided to select three inception blocks. In the last five images, there exist many objects with different scales and they are blended with each other, which makes them need four inception blocks.



	%
	\section{Conclusion}
	\label{Conclusion}
	In this work, we demonstrate that the combination of multiple convolutions improves detection accuracy by expanding the receptive fields and enriching the spatial information in the feature pyramid. The large amounts of computational cost lead by multiple convolutions can also be saved for some easier images. Thus, we propose a novel DyFPN for object detection. Based on the input image, the dynamic gate in DyFPN adaptively determines whether to execute the multiple convolutions in the feature pyramid. DyFPN can largely reduce the computational cost while preserving high accuracy. Experiments on various backbones demonstrate the effectiveness of the proposed DyFPN.
	
	%
	
	\bibliography{egbib}
\end{document}